\documentclass[a4paper]{article}

\usepackage{INTERSPEECH2021}

\usepackage{graphicx}
\graphicspath{{fig/}}
\newcommand{\figsep}{\vspace*{-4pt}}
\newcommand{\tablesep}{\vspace*{-6pt}}

\usepackage{booktabs}
\usepackage{tabularx}
\usepackage{multirow}
\newcommand{\mytable}{
	\centering
	\renewcommand{\arraystretch}{1.1}
	}

\newcolumntype{C}{>{\centering\arraybackslash}X}
\newcolumntype{L}{>{\raggedright\arraybackslash}X}
\newcolumntype{R}{>{\raggedleft\arraybackslash}X}
\newcolumntype{P}[1]{>{\raggedright\arraybackslash}p{#1}}
\usepackage{siunitx}
\usepackage{etoolbox}                           
\newcommand{\ubold}{\fontseries{b}\selectfont}  
\robustify\ubold                                

\usepackage{amsmath}
\usepackage{amssymb}
\DeclareMathOperator*{\argmin}{arg\,min}
\renewcommand{\vec}[1]{\boldsymbol{\mathbf{#1}}}

\usepackage{calc}
\usepackage{microtype}
\usepackage{url}
\usepackage{xcolor}

\usepackage{cite}
\let\oldbibliography\thebibliography
\renewcommand{\thebibliography}[1]{\oldbibliography{#1}
                                   \setlength{\itemsep}{0.2mm}
                                   \vspace*{-0mm}}

\title{Towards unsupervised phone and word segmentation \\ using self-supervised vector-quantized neural networks}
\name{Herman Kamper\sthanks{~~The two authors contributed equally.} \qquad Benjamin van Niekerk\footnotemark[1]}
\address{Department of E\&E Engineering, Stellenbosch University, South Africa}
\email{kamperh@sun.ac.za, benjamin.l.van.niekerk@gmail.com}

\usepackage[prependcaption,textsize=scriptsize]{todonotes}
\definecolor{mycolor}{HTML}{FF6600}

\begin{document}

\maketitle

\begin{abstract}
We investigate segmenting and clustering speech into low-bitrate phone-like sequences without supervision. We specifically constrain pretrained self-supervised vector-quantized~(VQ) neural networks so that blocks of contiguous feature vectors are assigned to the same code, thereby giving a variable-rate segmentation of the speech into discrete units. Two segmentation methods are considered. In the first, features are greedily merged until a prespecified number of segments are reached. The second uses dynamic programming to optimize a squared error with a penalty term to encourage fewer but longer segments. We show that these VQ segmentation methods can be used without alteration across a wide range of tasks: unsupervised phone segmentation, ABX phone discrimination, same-different word discrimination, and as inputs to a symbolic word segmentation algorithm. The penalized dynamic programming method generally performs best. While performance on individual tasks is only comparable to the state-of-the-art in some cases, in all tasks a reasonable competing approach is outperformed at a substantially lower bitrate.
\end{abstract}

\section{Introduction}

Methods for automatically learning phone- or word-like units from unlabelled speech audio could enable speech technology in severely low-resourced settings~\cite{jansen+etal_icassp13,menon+etal_interspeech19} and could lead to new cognitive models of human language acquisition~\cite{rasanen_speechcom12,dupoux_cognition18,shain+elsner_conll20}.
The goal in unsupervised representation learning of phone units is to learn features which capture phonetic contrasts while being invariant to properties like the speaker or channel.
Early approaches focussed on learning continuous features~\cite{zeghidour+etal_interspeech16,heck+etal_ieice18,chung+etal_interspeech19,last+etal_spl20,algayres+etal_interspeech20}.
In an attempt to better match the categorical nature of true phonetic units, more recent work has considered \textit{discrete} representations~\cite{badino+etal_interspeech15,chorowski+etal_taslp19,dunbar+etal_interspeech19,eloff+etal_interspeech19,baevski+etal_iclr20,chen+hain_interspeech20,chung+etal_interspeech20}.
One approach is to use a self-supervised neural network with an intermediate layer that quantizes features using a learned codebook~\cite{chorowski+etal_taslp19}.
While the discrete codes from such vector quantized (VQ) networks have given improvements in intrinsic phone discrimination tasks, they still encode speech at a much higher bitrate than true phone sequences~\cite{dunbar+etal_interspeech20}.

\begin{figure}[!b]
	\centering
	\includegraphics[width=0.99\linewidth]{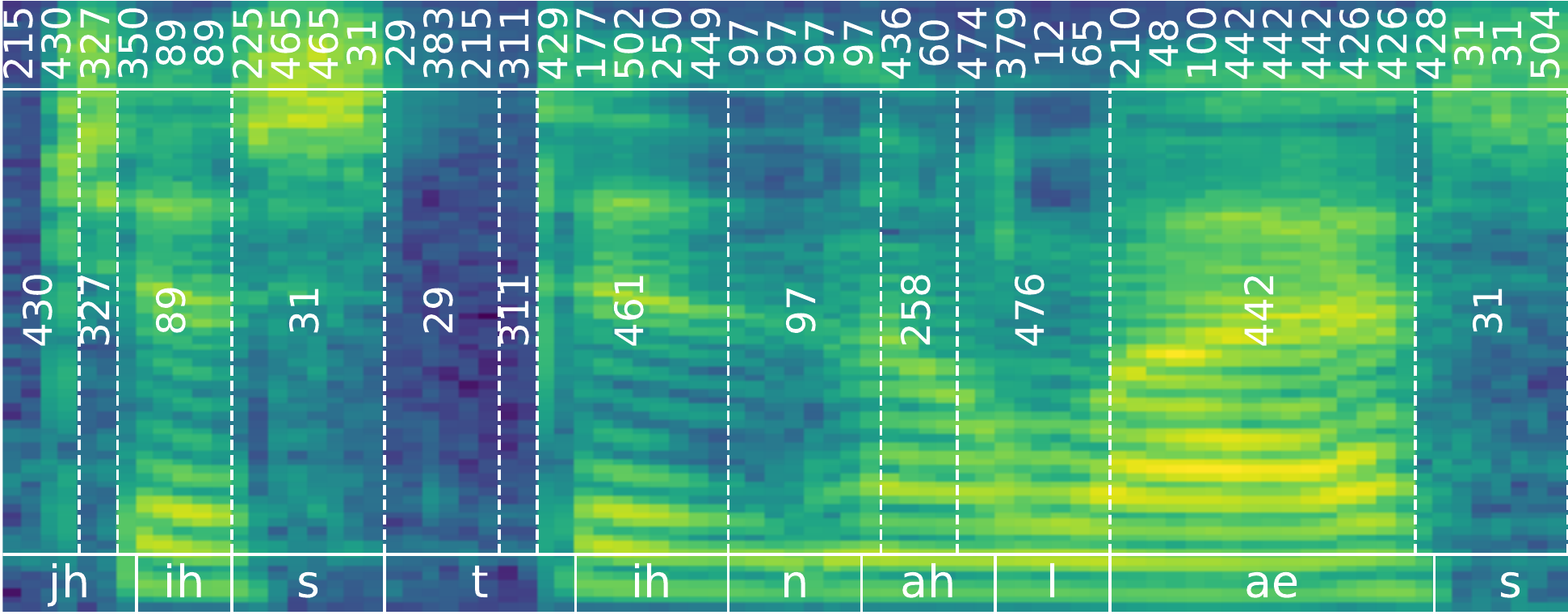}
	\figsep
	\caption{The phrase `just in the last' 
		with the code indices from a VQ-VAE model (top) and the ground truth phones (bottom). The indices with the dashed segmentation (middle) is the result of applying duration-penalized dynamic programming segmentation on the VQ-VAE codes.}
	\label{fig:vqvae_output}
\end{figure}

As an example, the top of Figure~\ref{fig:vqvae_output} shows the code indices from a vector-quantized variational autoencoder (VQ-VAE)~\cite{vandenoord+etal_neurips17} overlaid on the input spectrogram.
While there is some correspondence between the code assignments and the true phones (e.g.\ code 31 in both occurrences of [s]), and although there is some repetition of codes in adjacent frames (e.g.\ in [n] and [ae]), the input speech are often assigned to codes that are distinct from those of surrounding frames.
This is not surprising since the VQ model is not explicitly encouraged to do so.
The result is an encoding at a much higher bitrate (around 400 bits/sec) than that of true phone sequences (which is around 40 bits/sec~\cite{coupe+etal_sciadv19,dunbar+etal_interspeech20}).

In this paper we consider ways to constrain VQ models so that contiguous feature vectors are assigned to the same code, resulting in a low-bitrate segmentation of the speech into discrete units.
We specifically compare two VQ segmentation methods.
Both of these are based on a recent method for segmenting written character sequences~\cite{chorowski+etal_neuripspgr9}.
The first method is a greedy approach, where the closest adjacent codes are merged until a set number of segments are reached.
The second method allows for an arbitrary number of segments. A squared error between blocks of feature vectors and VQ codes are used together with a penalty term encouraging longer-duration segments. The optimal segmentation is found using dynamic programming.

We apply these two segmentation approaches using the encoders and codebooks of the two VQ models from \cite{vanniekerk+etal_interspeech20}.
The first is a type of VQ-VAE.
The second is a vector-quantized contrastive predictive coding (VQ-CPC) model.
The combination of these two models with the two segmentation approaches gives a total of four VQ segmentation models to consider.
We evaluate these on
four different tasks: unsupervised phone segmentation~\cite{rasanen_cogsci14}, ABX phone discrimination~\cite{schatz+etal_interspeech13}, same-different word discrimination~\cite{carlin+etal_icassp11}, and as inputs to a symbolic word segmentation algorithm~\cite{jansen+etal_icassp13}.
The last-mentioned is particularly important since the segmentation and clustering of 
word-like units 
remains a major but important challenge~\cite{shain+elsner_conll20,rasanen+blandon_interspeech20}.

On most metrics in the four tasks the combination of the VQ-VAE with the penalized dynamic programming approach is the best VQ segmentation method.
Example output is shown in the middle of Figure~\ref{fig:vqvae_output}.
Compared to other existing methods, it does not achieve state-of-the-art performance in all four evaluation tasks.
However, it achieves reasonable performance at a much lower bitrate than most existing methods.
This is noteworthy since, while most of the other methods have been tailored to the respective tasks, a single VQ segmentation approach can be used without any alteration directly in a range of problems.

\section{Vector-quantized neural networks} 
\label{sec:vqvae_vqcpc}

A vector quantization (VQ) layer~\cite{vandenoord+etal_neurips17} consists of a trainable codebook $\{\vec{e}_1, \vec{e}_2, \ldots, \vec{e}_K \}$ of $K$ distinct codes.
In the forward pass, the layer discretizes a sequence of continuous feature vectors $(\vec{z}_1, \vec{z}_2, \ldots, \vec{z}_T)$ by mapping each $\vec{z}_t$ to it's nearest neighbour in the codebook, i.e.\ each $\vec{z}_t$ is replaced with $\vec{e}_k$, where $k = \argmin_i || \vec{z}_t - \vec{e}_i ||^2$.
The layer output is the resulting quantized sequence $(\hat{\vec{z}}_1, \hat{\vec{z}}_2, \ldots, \hat{\vec{z}}_T)$.
Since the $\argmin$ operator is not differentiable, gradients are approximated using the straight-through estimator \cite{bengio+etal_arxiv13} in the backward pass. 
The codebook is trained using an exponential moving average of the continuous features.
A commitment cost is also added to encourage each $\vec{z}_t$ to commit to its selected code.

In this paper we use the pretrained encoders and codebooks of the two self-supervised VQ models from~\cite{vanniekerk+etal_interspeech20}.
Both models' encoders take log-Mel spectrograms, downsamples it by a factor of two, and discretizes the feature vectors using a VQ layer with 512 codes.
The first model is a type of vector-quantized variational autoencoder (VQ-VAE).\footnote{\scriptsize \url{https://github.com/bshall/ZeroSpeech}}
The VQ-VAE is trained to encode speech into a discrete latent space from which it reconstructs the original audio waveform using an autoregressive decoder~\cite{chorowski+etal_taslp19}.
The second model\footnote{\scriptsize \url{https://github.com/bshall/VectorQuantizedCPC}} combines vector quantization with contrastive predictive coding (VQ-CPC), similar to the models of~\cite{baevski+etal_iclr20,hadjeres+crestel_arxiv20}.
Using a contrastive loss~\cite{vandenoord+etal_arxiv18}, the VQ-CPC model is trained to identify future codes from among a set of negative examples drawn from other utterances.

\section{VQ segmentation algorithms}
\label{sec:vqseg}

\begin{figure}[!t]
	\centering
	\includegraphics[width=0.9\linewidth]{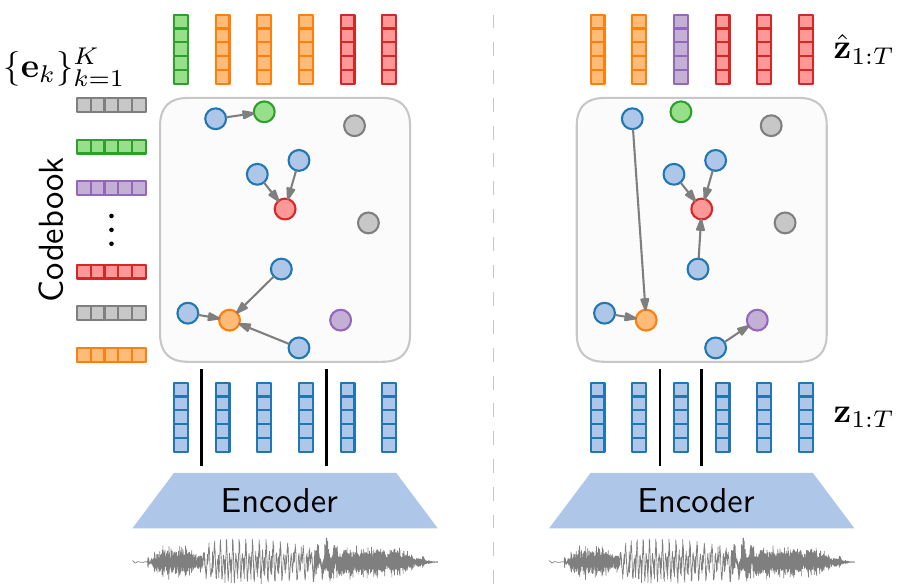}
	\figsep
	\caption{Two potential segmentations of an utterance.
	The features in each segment are assigned to the same code.
	The left segmentation will have a smaller sum of (squared) distances between the features and their assigned codes than the right.
	}
    \vspace*{-3pt}
	\label{fig:vq_segmentation_example}
\end{figure}

The segmentation process consists of two steps. 
First, we extract a sequence of continuous feature vectors using a pretrained encoder.
Next, we solve a constrained optimization problem to divide the continuous representation into segments.
Figure~\ref{fig:vq_segmentation_example} illustrates the steps, showing two possible segmentations.
Each segment is assigned a representative vector chosen from the codebook.
We can score the segmentations by summing the (squared) distances between the continuous feature vectors and the representative code of each segment (this corresponds to adding up the lengths of the arrowed lines in Figure~\ref{fig:vq_segmentation_example}).
The objective would then be to find the segmentation that minimizes the summed distance.
One problem with blindly following this approach is that the best segmentation will always put each $\vec{z}_t$ in its own segment, assigning it to the code closest to $\vec{z}_t$.
This would just be the standard VQ layer~(\S\ref{sec:vqvae_vqcpc}).
Constraints are therefore required.

Formally, suppose sequence $(\mathbf{z}_1, \mathbf{z}_2, \ldots, \mathbf{z}_T)$ is segmented into disjoint contiguous segments $(s_1, s_2, \ldots, s_M)$. Each segment $s_i$ is a subsequence of $\mathbf{z}_{1:T}$, consisting of $|s_i|$ feature vectors.
These vectors are aggregated and the
whole segment $s_i$ is assigned to a single representative vector $\hat{\mathbf{z}}_{s_i}$, selected from the VQ codebook $\{ \mathbf{e}_k \}_{k = 1}^K$.
So, as a concrete example using this notation, we might have $s_2 = \mathbf{z}_{3:5}$ with $\hat{\mathbf{z}}_{s_2} = \mathbf{e}_7$, meaning that all three vectors ($|s_2| = 3$) in the second segment are assigned to the seventh code.

Let's use $E(\mathbf{z}_{1:T}, s_{1:M})$ to denote the error for a particular
segmentation $s_{1:M}$ of the features $\mathbf{z}_{1:T}$. One option for
the error function could be
\begin{equation}
	E(\mathbf{z}_{1:T}, s_{1:M}) = \sum_{s_i \in s_{1:M}} \sum_{\mathbf{z}_j \in s_i} || \mathbf{z}_j - \hat{\mathbf{z}}_{s_i} ||^2.
	\label{eq:squared}
\end{equation}
The problem with this error function, as noted above, is that the optimal
strategy would simply be to place each $\mathbf{z}_t$ in its own segment. The
solution is to add a penalty term that encourages longer but fewer segments:
\begin{equation}
	E(\mathbf{z}_{1:T}, s_{1:M}) = \sum_{s_i \in s_{1:M}} \sum_{\mathbf{z}_j \in s_i} \left[ || \mathbf{z}_j - \hat{\mathbf{z}}_{s_i} ||^2 + \lambda \, \textrm{pen}(|s_j|) \right],
	\label{eq:dp_penalized}
\end{equation}
where $\textrm{pen}(|s_j|)$ is a function penalizing the number of frames $|s_j|$ in a segment and $\lambda$ is the penalty weight.
Our objective is then
to find the optimal segmentation $\argmin_{s_{1:M}}
E(\mathbf{z}_{1:T}, s_{1:M})$, and this can be accomplished using dynamic
programming. Concretely, we define forward variables
$
\alpha_t \triangleq \min_{s_{1:M_t}} E(\mathbf{z}_{1:t}, s_{1:M_t})
$
as the error for the optimal segmentation up to step $t$. This can be
calculated recursively:
$$
\alpha_t = \min_{j = 1}^t \Big\{ \alpha_{t - j} + \min_{k = 1}^K \mkern-5mu \sum_{\mathbf{z}_i \in \mathbf{z}_{t-j+1:t}} \mkern-20mu \left[ || \mathbf{z}_i - \mathbf{e}_k ||^2  + \lambda \, \textrm{pen} (j) \right] \Big\}.
$$
We start with $\alpha_0 = 0$ and calculate $\alpha_t$ for $t = 1, \ldots, T -
1$. We keep track of the optimal choice ($\argmin$) for each
$\alpha_t$, and the overall optimal segmentation is then obtained by starting
from the final position $t = T$ and moving backwards, repeatedly choosing the
optimal boundary.

The above shortest-path (i.e.\ Viterbi) formulation is a generalization of the approach of~\cite{chorowski+etal_neuripspgr9}. Their starting point for the constraint on~\eqref{eq:squared} is different,
however. Instead of using a penalty term, they enforce a prespecified number of
segments. I.e., while above $M$ can take on an arbitrary number (depending on
the penalty weight $\lambda$), they require a specific number of segments:
\begin{equation}
	E(\mathbf{z}_{1:T}, s_{1:M}) = \sum_{s_i \in s_{1:M}} \sum_{\mathbf{z}_j \in s_i} || \mathbf{z}_j - \hat{\mathbf{z}}_{s_i} ||^2 \textrm{\ \ s.t.\ } M = N,
	\label{eq:chorowski}
\end{equation}
where $N$ is the set number of segments.

Actually, if we set the penalty function in~\eqref{eq:dp_penalized} to $\textrm{pen}(|s_j|) = 1 - |s_j|$, then it can be shown that the formulation in~\eqref{eq:dp_penalized} is exactly the
dual of~\eqref{eq:chorowski}. 
In our experiments we indeed use this penalty function---it gave reasonable results on development data and we did not experiment further.
A more fundamental implementational difference here is that~\cite{chorowski+etal_neuripspgr9} doesn't actually optimise~\eqref{eq:chorowski} directly.
Instead they use an approximation where adjacent vectors are greedily merged until exactly $N$ segments are obtained. The greedy approach has the advantage that a solution can be found in $\mathcal{O}(KT \log T)$ time,\footnote{Our version of the greedy approach is slower than this, though, since we implement it with the help of an agglomerative clustering package.} while getting the exact solution using full dynamic programming takes $\mathcal{O}(KT^2N)$ time.

\section{Experimental setup}
\label{sec:experimental_setup}

Both the VQ-VAE and VQ-CPC models (\S\ref{sec:vqvae_vqcpc})
are trained on the English training set from the \textit{ZeroSpeech 2019 Challenge}~\cite{dunbar+etal_interspeech19}, consisting of around 15 hours of speech from over 100 speakers.
Depending on the task, evaluation is performed on the English test set from \textit{ZeroSpeech 2019}, with around 30 minutes of speech, or on the Buckeye English development or test sets~\cite{pitt+etal_speechcom05}, each roughly 6 hours. 
There is no speaker overlap between these sets and the one used for training.

Hyperparameters for the VQ segmentation methods (\S\ref{sec:vqseg}) were tuned on Buckeye development data.
The dynamic programming penalized method (denoted as `DP penalized' in the tables and figures below) has one hyperparameter, $\lambda$, the duration penalty weight in~\eqref{eq:dp_penalized}.
We set $\lambda = 3$ for the VQ-VAE and to $\lambda = 400$ for the VQ-CPC.
For the greedy $N$-segmentation method (denoted as `Greedy $N$-seg.' below), 
we need to specify the number of segments $N$ that each utterance is segmented into.
Instead of setting $N$ directly, we rather set the required number of frames per segment: 3 gave the best development results.

We have four model combinations (VQ-VAE/VQ-CPC with greedy/DP penalized segmentation), which we evaluate on four different tasks.
For the phone and word segmentation tasks below, we measure precision, recall and $F$-score of boundaries with a tolerance of 20~ms.
We also measure over-segmentation~(OS): how many more/fewer boundaries are proposed than the ground truth, e.g.\ an OS of --0.1 indicates we have 10\% too few boundaries.
$F$-score isn't always sensitive enough to the trade-off between recall and OS, which motivated the $R$-value metric~\cite{rasanen+etal_interspeech09}: it gives a perfect score (1) when a method has perfect recall (1) and perfect OS~(0).

\section{Experiments}

\subsection{Task 1: Phone segmentation}

We perform phone segmentation experiments on Buckeye following the setup of~\cite{kreuk+etal_interspeech20}. 
In Table~\ref{tbl:phoneseg_dev} we first compare the different VQ segmentation approaches to each other on development data.
For the `Merged' rows, repeated codes from the VQ models are simply collapsed.
While this gives high recall, this is because of very high OS.
Of the two VQ segmentation approaches, the dynamic programming variant (DP penalized) gives better results than the greedy version (Greedy $N$-seg.) in all cases except in OS on VQ-CPC codes.
Using the VQ-VAE encoder and codebook is consistently better than using that of the VQ-CPC.

\begin{table}[!b]
	\mytable
	\caption{A comparison of different VQ segmentation algorithms for phone segmentation (\%) on Buckeye development data.
		Segmentation of both VQ-CPC and VQ-VAE codes are shown.}
	\eightpt
	\tablesep
	\begin{tabularx}{\linewidth}{@{}LlcccS[table-format=2.1]S[table-format=2.1]@{}}
		\toprule
		Model & VQ seg. & Prec. & Rec. & $F$ & OS & {$R$-val.} \\
		\midrule
		
		\textit{\underline{VQ-CPC:}} & 
		Merged & 29.7 & \ubold 98.9 & 45.7 & 232.6 & -98.9 \\
		& Greedy $N$-seg. & 51.5 & 65.6 & 57.7 & 27.4 & 56.2 \\
		& DP penalized & 55.7 & 74.7 & 63.8 & 34.0 & 57.8 \\
		\addlinespace
		
		\textit{\underline{VQ-VAE:}} & 
		Merged & 32.0 & 98.3 & 48.3 & 207.2 & -77.4 \\
		& Greedy $N$-seg. & 57.1 & 72.9 & 64.0 & 27.8 & 61.2 \\
		& DP penalized & \ubold 66.4 & 75.8 & \ubold 70.8 & \ubold 14.1 & \ubold 72.4 \\				
		\bottomrule
	\end{tabularx}
	\label{tbl:phoneseg_dev}
\end{table}

For testing, we therefore use the VQ-VAE with DP penalized segmentation.
This is also the method in the middle of Figure~\ref{fig:vqvae_output}.
Table~\ref{tbl:phoneseg_test} compares this approach to existing methods.
While our unsupervised approach does not perform as well as~\cite{kreuk+etal_interspeech20} on all metrics, it performs similarly or better than~\cite{michel+etal_arxiv16} and~\cite{wang+etal_interspeech17}.

\begin{table}[!b]
	\mytable
	\caption{Phone segmentation results (\%) on Buckeye test data for existing methods and VQ segmentation.}
	\eightpt
	\tablesep
	\begin{tabularx}{\linewidth}{@{}lCCCS[table-format=2.1]S[table-format=2.1]@{}}
		\toprule
		Model & Prec. & Rec. & $F$ & {OS} & {$R$-val.} \\
        \midrule
		\textit{\underline{Unsupervised:}} \\[2pt]
		Next-frame prediction~\cite{michel+etal_arxiv16} & 69.3 & 65.1 & 67.2 & -6.1 & 72.1 \\
		GRU gate activation~\cite{wang+etal_interspeech17} & 69.6 & 72.6 & 71.0 & -4.1 & 74.8 \\
		Self-sup.\ contrastive~\cite{kreuk+etal_interspeech20} &  \ubold 75.8 & 76.9 & 76.3 & \ubold -1.4 &  \ubold 79.7 \\[3pt]
		Our VQ-VAE: DP penalized & 70.8 & \ubold 85.6 &  \ubold 77.5 & 20.9 & 74.8 \\
		
		\addlinespace
		\textit{\underline{Supervised:}} \\[2pt]
		LSTM~\cite{franke+etal_itg16} & 87.8 & 83.3 & 85.5 & 5.4 & 87.2 \\
		LSTM structured loss~\cite{kreuk+etal_icassp20} & 85.4 & 89.1 & 87.2 & -4.1 & 88.8 \\
		\bottomrule
	\end{tabularx}
	\label{tbl:phoneseg_test}
\end{table}

\begin{table}[!b]
	\mytable
	\caption{ABX phone discrimination error rates and bitrates (bits/sec) on the ZeroSpeech 2019 English data.}
	\eightpt
	\tablesep
	\begin{tabularx}{\linewidth}{@{}LcS[table-format=4.0]@{}}
		\toprule
		Model & ABX (\%) & {Bitrate} \\
		\midrule	
		\textit{\underline{Unsupervised:}} \\[2pt]
		MFCCs~\cite{eloff+etal_interspeech19} & 22.7 & 1738 \\[3pt]
		Instance normalized WaveNet AE~\cite{chen+hain_interspeech20} & 20.2 & 386 \\
		VQ-CPC~\cite{vanniekerk+etal_interspeech20} & \ubold 13.4 & 421 \\
		VQ-VAE~\cite{vanniekerk+etal_interspeech20} & 14.0 & 412 \\[3pt]
		Our VQ-VAE: Greedy $N$-seg. & 23.1 & 142 \\
		Our VQ-VAE: Duration penalized & 18.5 & \ubold 106 \\	
		\addlinespace
		
		\textit{\underline{Supervised:}} \\[2pt]
		ASR output~\cite{dunbar+etal_interspeech20} & 29.9 & 38 \\
		\bottomrule
	\end{tabularx}
	\label{tbl:abx}
\end{table}

\subsection{Task 2: ABX phone discrimination}

Apart from segmentation performance, we want to see whether the resulting discrete sequences retain phonetic information, and whether it can do so at a low bitrate.
For this we use the ABX phone discrimination task~\cite{schatz+etal_interspeech13}.
This tests whether triphone $X$ is more similar to triphones $A$ or $B$, where $A$ and $X$ are instances of the same triphone (e.g.\ `cat'), while $B$ differs in the middle phone (e.g.\ `cut').
To measure speaker-invariance, $A$ and $B$ come from the same speaker, while $X$ is taken from a different speaker.
As a similarity metric, we use the average cosine distance along the dynamic time warping (DTW) alignment path over the segmented codes.
ABX is reported as an aggregated error rate over all pairs of triphones in the test set.

Table~\ref{tbl:abx} compares our approaches to some of the best submissions to \textit{ZeroSpeech 2020} (a repeat of the 2019 challenge): the submission of~\cite{vanniekerk+etal_interspeech20} placed first on ABX followed by~\cite{chen+hain_interspeech20}.
While VQ segmentation gives worse ABX than the unconstrained VQ-CPC and VQ-VAE, the bitrate is also dramatically reduced.
Moreover, despite the drop, DP penalized segmentation still achieves better ABX than the closest competitor~\cite{chen+hain_interspeech20}
(18.5\% vs 20.2\%) at roughly a third of the bitrate (106~vs~386).

\begin{figure}[!t]
	\centering
	\includegraphics[width=0.99\linewidth]{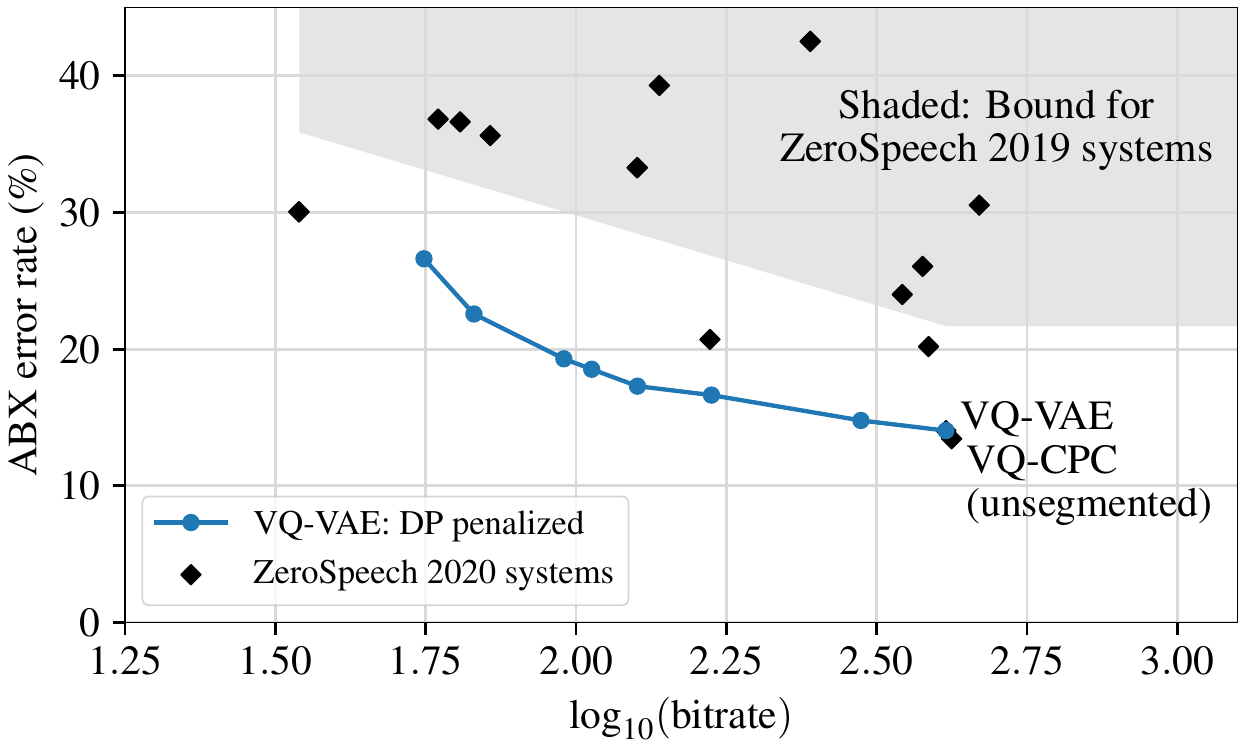}
	\figsep
	\caption{ABX error rate as a function of bitrate on the ZeroSpeech 2019 English data. For the DP penalized method, bitrates are varied by changing the duration penalty weight $\lambda$.}
	\label{fig:abx}
\end{figure}

An advantage of VQ segmentation is that the degree of compression can be controlled.
Figure~\ref{fig:abx} illustrates the trade-off between phone discrimination performance and compression as the penalty weight parameter $\lambda$ is varied~(\S\ref{sec:experimental_setup}).
The 
DP penalized VQ-VAE comes closer to the ideal of a 0\% error rate at a 0 bitrate than submissions to either \textit{ZeroSpeech 2019} or \textit{2020}.

\subsection{Task 3: Same-different word discrimination}

\begin{table}[!t]
	\mytable
	\caption{Average precision (AP, \%) and bitrates (bits/sec) achieved in same-different word discrimination on Buckeye data.
	Only a selection of models are applied to the test data.}
	\eightpt
	\tablesep
	\begin{tabularx}{\linewidth}{@{}LcS[table-format=4.0]cS[table-format=4.0]@{}}
		\toprule
		& \multicolumn{2}{c}{Development} & \multicolumn{2}{c}{Test} \\
		\cmidrule(l){2-3} \cmidrule(l){4-5}
		Model & AP & {Bitrate} & AP & {Bitrate} \\
		\midrule
		MFCCs & 36.8 & 1721 & 35.9 & 1746 \\[3pt]
		CAE~\cite{renshaw+etal_interspeech15} & 47.4 & 1721 \\
		Siamese~\cite{synnaeve+etal_slt14,zeghidour+etal_icassp16} & 38.2 & 1721 \\
		CTriamese~\cite{last+etal_spl20} & \ubold 50.4 & 1721 & 47.2 & 1746 \\
		VQ-CPC~\cite{vanniekerk+etal_interspeech20} & 45.6 & 429  \\
		VQ-VAE~\cite{vanniekerk+etal_interspeech20} & \ubold 50.4 & 410 & \ubold 47.8 & 411 \\[3pt]
		Our VQ-VAE: Merged & 47.9 & 310 \\  
		Our VQ-VAE: Greedy $N$-seg. & 25.7 & 135 \\  
		Our VQ-VAE: DP penalized & 35.7 & \ubold 118 & 34.5 & \ubold 118 \\
		
		\bottomrule
	\end{tabularx}
	\label{tbl:same_different}
\end{table}

In the same-different word discrimination task~\cite{carlin+etal_icassp11}, we are given a pair of spoken words and we must decide whether they are examples of the same or different words.
For every word pair in a test set of pre-segmented words, the DTW distance is calculated using the feature representation under evaluation. Two words can then be classified as being of the same or different type 
based on some threshold, and a precision-recall curve is obtained by varying the threshold. The area under this curve is used as final evaluation metric, referred to as the average precision (AP).

Table~\ref{tbl:same_different} compares our approaches to existing unsupervised methods.
We again see that VQ segmentation trades off word discrimination performance for compression, e.g.\ the DP penalized VQ-VAE achieves a similar AP to MFCCs at less than a tenth of the bitrate.
Also note that the results for the unsegmented VQ-CPC and VQ-VAE given here are new---they have never been applied to this task.
This is therefore the first time that it is shown that the unsegmented
VQ-VAE achieve state-of-the-art word discrimination scores (47.8\% on test) at a much lower bitrate than the current best approach~\cite{last+etal_spl20} (411 vs 1746 on test).

\subsection{Task 4: Towards word segmentation}

Inspired by~\cite{jansen+etal_icassp13}, we consider how word segmentation can be performed on top of VQ segmentation: code indices are provided as input to a symbolic word segmentation algorithm, normally used for segmenting phonemic or character sequences without word spaces.
We use the adaptor grammar (AG)~\cite{johnson+etal_nips06}.
Since this is an initial experiment, 
we report development~results. 
We also performed initial experiments with the unigram Dirichlet process model of~\cite{goldwater+etal_cognition09} and a method based on thresholding the transition probabilities between code indices~\cite{saksida+etal_develsci17}. The thresholding approach performed worst while the Dirichlet process and AG performed similarly. We only report results for the latter.

Table~\ref{tbl:wordseg_val} shows word boundary segmentation scores for existing approaches (top); VQ segmentation where the code sequences are treated as words themselves (middle); and when the VQ segmentation code indices are segmented with an AG (bottom).
While the VQ segmentation approaches (middle) achieve higher word boundary $F$-scores, they heavily over-segment, achieving much worse $R$-value scores than previous methods (top).
When segmenting the code indices with an AG (bottom), we see that OS does improve compared to just treating the segmented codes as words (middle). 
However, our best approach (AG on top of VQ-CPC DP penalized indices) still heavily over-segments.
Word token $F$-scores, where both the beginning and end boundaries need to be correctly predicted without an intermediate boundary, is very poor (in the order of 4\%, not shown in the table). This is similar to the results in~\cite{jansen+etal_icassp13} (which was performed on a different dataset).
As in~\cite{jansen+etal_icassp13}, we conclude that a more integrated approach is required where word segmentation is (potentially) performed jointly with VQ segmentation.

\begin{table}[!t]
	\mytable
	\caption{Word boundary segmentation results (\%) on Buckeye development data for existing methods and VQ segmentation code indices segmented with an adaptor grammar.}
	\eightpt
	\tablesep
	\begin{tabularx}{\linewidth}{@{}LcccS[table-format=2.1]S[table-format=2.1]@{}}
		\toprule
		Model & Prec. & Rec. & $F$ & {OS} & {$R$-val.} \\
		\midrule
		\textit{\underline{Existing methods:}} \\[2pt]
		ES-KMeans~\cite{kamper+etal_asru17} & 30.7 & 18.0  & 22.7 & \ubold -41.2 & \ubold 39.7 \\
		BES-GMM~\cite{kamper+etal_csl17} & \ubold 31.7 & 13.8 & 19.2 & -56.6 & 37.9 \\
		
		\addlinespace
		\multicolumn{3}{@{}l}{\textit{\underline{Treating VQ seg.\ as word seg.:}}} \\[2pt]
		VQ-CPC: DP penalized & 15.5 & \ubold 81.0 & 26.1 & 421.4 & -266.6 \\
		VQ-VAE: DP penalized & 15.8 & 68.1 & 25.7 & 330.9 & -194.5 \\
	
		\addlinespace
		\multicolumn{3}{@{}l}{\textit{\underline{Adaptor grammar on top of:}}} \\[2pt]
		VQ-CPC: DP penalized & 18.2 & 54.1 & \ubold 27.3 & 196.4 & -86.5 \\ 
		VQ-VAE: DP penalized & 16.4 & 56.8 & 25.5 & 245.2 & -126.5 \\
		\bottomrule
	\end{tabularx}
	\label{tbl:wordseg_val}
\end{table}

\section{Discussion and conclusion}

We considered methods for constraining the outputs of vector-quantized (VQ) neural networks to produce a discrete low-bitrate segmentation of unlabelled speech.
The experimental results on four different tasks show that our VQ segmentation method gives reasonable performance at a much lower bitrate than existing methods.
While we found that the segmented codes from a VQ-VAE is normally slightly better than that from a VQ-CPC, it is worth noting that the latter is typically faster to train, since it does not require waveform reconstruction.
Another practical consideration is that, while duration-penalized segmentation with dynamic programming typically outperformed the greedy approach where the number of segments are specified, the former has a more sensitive duration-weight hyperparameter, which could make the latter more useful in practice.
In future work, we would like to further investigate whether the resulting discrete units match true phonetic units. We would also like to see whether VQ segmentation can be performed as part of a VQ network, rather than using pretrained encoders and codebooks.

\vspace{3pt}
{\eightpt
\noindent \textbf{Acknowledgements.} We would like to thank Felix and Yossi$^2$ for sharing experimental details~\cite{kreuk+etal_interspeech20}. This work is supported by the National Research Foundation of South Africa (grant no.\ 120409), a Google Faculty Award, and a Google Africa PhD Fellowship.}

\bibliography{mybib}

\end{document}